\def\year{2022}\relax
\documentclass[letterpaper]{article} 
\usepackage{aaai22}  
\usepackage{times}  
\usepackage{helvet}  
\usepackage{courier}  
\usepackage[hyphens]{url}  
\usepackage{graphicx} 
\usepackage{natbib}  
\usepackage{caption} 
\DeclareCaptionStyle{ruled}{labelfont=normalfont,labelsep=colon,strut=off} 
\usepackage{algorithm}
\usepackage{algorithmic}

\usepackage{newfloat}
\usepackage{listings}
\floatstyle{ruled}
\newfloat{listing}{tb}{lst}{}
\floatname{listing}{Listing}

\usepackage[colorlinks,
linkcolor=black,
anchorcolor=black,
urlcolor=black,
citecolor=black]{hyperref} 
\usepackage{url}
\usepackage{amsmath}
\usepackage{booktabs}
\usepackage{threeparttable}
\usepackage{multirow}
\usepackage{amsfonts}
\usepackage{subfigure}
\usepackage{bm}
\usepackage{color}

\title{Regularizing End-to-End Speech Translation with Triangular \\ Decomposition Agreement}
\author{
    Yichao Du$^{\ddag}$, Zhirui Zhang$^\natural$, Weizhi Wang$^{\S}$, Boxing Chen$^\natural$, Jun Xie$^\natural$ and Tong Xu$^{\ddag}$\thanks{Corresponding author.}
}
\affiliations{
    $^\ddag$University of Science and Technology of China, Hefei, China\\
    $^\natural$Machine Intelligence Technology Lab, Alibaba DAMO Academy \ \ $^\S$Rutgers University, New Brunswick, USA \\
    $^\ddag$duyichao@mail.ustc.edu.cn \ $^\ddag$tongxu@ustc.edu.cn  \\ 
    $^{\natural}$zrustc11@gmail.com $^{\natural}$\{boxing.cbx, qingjing.xj\}@alibaba-inc.com \ $^{\S}$weizhi.wang@rutgers.edu

}

\usepackage{bibentry}

\begin{document}

\maketitle

\begin{abstract}

End-to-end speech-to-text translation~(E2E-ST) is becoming increasingly popular due to the potential of its less error propagation, lower latency, and fewer parameters.
Given the triplet training corpus $\langle speech, transcription, translation\rangle$, the conventional high-quality E2E-ST system leverages the $\langle speech, transcription\rangle$ pair to pre-train the model and then utilizes the $\langle speech, translation\rangle$ pair to optimize it further.
However, this process only involves two-tuple data at each stage, and this loose coupling fails to fully exploit the association between triplet data.
In this paper, we attempt to model the joint probability of transcription and translation based on the speech input to directly leverage such triplet data.
Based on that, we propose a novel regularization method for model training to improve the agreement of dual-path decomposition within triplet data, which should be equal in theory.
To achieve this goal, we introduce two Kullback-Leibler divergence regularization terms into the model training objective to reduce the mismatch between output probabilities of dual-path.
Then the well-trained model can be naturally transformed as the E2E-ST models by the pre-defined early stop tag.
Experiments on the MuST-C benchmark demonstrate that our proposed approach significantly outperforms state-of-the-art E2E-ST baselines on all 8 language pairs, while achieving better performance in the automatic speech recognition task.

\end{abstract}

\section{Introduction}
Speech-to-text translation (ST) processes speech signals in a source language and generates the text in a target language.
Traditional ST approaches cascade automatic speech recognition (ASR) and machine translation (MT)~\cite{Ney1999SpeechTC, Sperber2017NeuralLM, Zhang2019LatticeTF, IranzoSnchez2020DirectSM}.
With the rapid development of deep learning, the neural networks which are widely used in ASR and MT have been adapted to construct a new end-to-end speech-to-text translation (E2E-ST) paradigm~\cite{Liu2019EndtoEndST,Wang2020BridgingTG,Dong2021ListenUA}.
This approach aims to overcome known limitations of the cascade one and learns a single unified encoder-decoder model, which is easier to deploy, has lower latency and has less error propagation.

\begin{figure}[t]
    \centering
    \includegraphics[width=0.98\columnwidth]{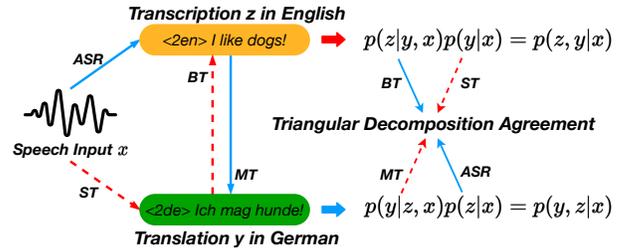}
    \caption{ A training exemplar of English-to-German speech-to-text task based on triangular decomposition. We can follow ASR-MT~(in blue line) and ST-BT~(in red line) decoding paths respectively to model the joint probability of transcription and translation given the speech input. According to the chain decomposition rule, the probability distribution of these two ways is equivalent.}
    \label{fig:intro_fig}
\end{figure}

Despite the potential benefits, it is very challenging to develop a well-trained E2E-ST model that does not use intermediate transcriptions.
Thus, various techniques have been proposed to ease the training process by using source transcriptions, including pre-training~\cite{Bansal2019PretrainingOH, Wang2020BridgingTG,Wang2020CurriculumPF}, multi-task learning~\cite{Anastasopoulos2018TiedML,Sperber2019AttentionPassingMF}, meta-learning~\cite{Indurthi2020EndendST}, consecutive decoding~\cite{Dong2021ConsecutiveDF}, and interactive decoding~\cite{Liu2020SynchronousSR}.
Among them, the pre-training strategy is a simple yet effective way, which is widely used to build the high-quality E2E-ST system in practice.
Specifically, given the triplet training dataset $\langle speech, transcription, translation\rangle$, the pre-training method first leverages the $\langle speech, transcription\rangle$ pair to pre-train the E2E-ST model via ASR and then further fine-tunes it with the $\langle speech, translation\rangle$ pair.
However, these methods only utilize two-tuple data at each stage, failing to fully explore the association relationship in triplet data. 
Therefore, how to fully mine the associations between the triplet data still remains a crucial issue to improve the performance of the E2E-ST model.
 
In this paper, we attempt to directly learn the joint probability of transcription and translation by a single unified encoder-decoder model and successive decoding to involve the whole triplet data.
Actually, as shown in Figure~\ref{fig:intro_fig}, there are two different paths in the decoding process to fit this joint probability according to the triangular decomposition:
(a) The speech signal $x$ is first converted into source transcription $z$ and then translated into target translation $y$. 
We name this decoding process as ASR-MT path, which can be formalized as $p(y,z|x)=p(z|x)p(y|z,x)$; 
(b) Meanwhile, we can directly perform speech-to-text translation on the speech signal and then adopt back-translation (BT) process to obtain the source transcription. 
This decoding process is formalized into $p(y,z|x)=p(y|x)p(z|y,x)$ and named as ST-BT path.
We adopt the language tag to distinguish these two different decoding paths.
Theoretically, the conditional distributions of such dual-path should be consistent due to the chain decomposition rule.
However, there is no guarantee that the above relationship will hold, if these two lines are learned separately.

Based on this observation, we propose a novel model regularization method called \textbf{T}riangular \textbf{D}ecomposition \textbf{A}greement (\textbf{TDA}) to better exploit the association between triplet data.
This goal is achieved by introducing two Kullback-Leibler divergence regularization terms into the original training objective to enhance the agreement between output probabilities of dual-path.
In this way, we not only maximize the probability distribution of the triplet training data, but also minimize the mismatch between ASR-MT and ST-BT decoding paths to promote the training process in the correct direction. 
In the inference stage, the well-trained model can be naturally transformed as the ASR and E2E-ST models by choosing ASR-MT and ST-BT decoding paths respectively, thus keeping the same inference delay as the previous ASR and E2E-ST models.

Our experiments are conducted on the MuST-C benchmark with all 8 language pairs, and demonstrate that our proposed approach not only gains up to 1.8 BLEU score improvements over the E2E-ST baseline on average but also achieves better performance in the ASR task. Our code is open-sourced at \url{https://github.com/duyichao/E2E-ST-TDA}.

\section{Background: End-to-End Speech Translation}
In this section, we first give a formal definition of the E2E-ST task, then briefly introduce the backbone model we use.
\paragraph{Problem Formulation.}
The ST training corpus consists of a set of triplet
data $\mathcal{D}_{ST}=\left\{(\mathbf{x}^{(n)}, \mathbf{z}^{(n)}, \mathbf{y}^{(n)})\right\}_{n=1}^{N}$.
Here $\mathbf{x}^{(n)}=(x_1^{(n)}, x_2^{(n)},...,x_{|\mathbf{x}^{(n)}|}^{(n)})$ denotes the input sequence of the speech wave (in most cases, acoustic features are used), $\mathbf{z}^{(n)}=(z_1^{(n)},z_2^{(n)},...,z_{|\mathbf{z}^{(n)}|}^{(n)})$ is the transcription sequence from the source language and the $\mathbf{y^{(n)}}=(y_1^{(n)}, y_2^{(n)},...,y_{|\mathbf{y}^{(n)}|}^{(n)})$ represents the translation sequence of target language. 
The goal of E2E-ST is to directly seek an optimal translation sequence $\mathbf{y}$ without generating an intermediate transcription $\mathbf{z}$, and the standard training objective is to optimize the maximum likelihood estimation (MLE) loss of the training data:
\begin{equation}
    \mathcal{L}_{MLE}(\theta)=\sum_{n=1}^{N} \log P\left(\mathbf{y^{(n)}}  \mid \mathbf{x^{(n)}} ; \theta\right),
    \label{eq:vanilla_st}
\end{equation}
where we use a single encoder-decoder structure to learn the conditional distribution $P(\mathbf{y^{(n)}}|\mathbf{x^{(n)}})$ and $\theta$ is the model parameter. 
In order to obtain the high-quality E2E-ST system, previous methods usually leverage ASR and MT tasks ($\{\mathbf{x}^{(n)}, \mathbf{z}^{(n)}\}$ and $\{\mathbf{z}^{(n)}, \mathbf{y}^{(n)}\}$) to pre-train the encoder and decoder respectively~\cite{Bansal2019PretrainingOH, Wang2020BridgingTG, Wang2020CurriculumPF}.
Since this process only adopts two-tuple data at each stage, this loose coupling fails to fully utilize the association between triplet data.
 
\paragraph{Backbone E2E-ST Model.}
In this work, we adopt the transformer-based~\cite{Vaswani2017AttentionIA} structure as the backbone, which has become increasingly common in the speech processing field.
Concretely, the entire encoder consists of a multi-layer convolutional down-sampling module and a transformer-based encoder. 
The multi-layer convolutional module takes the acoustic features as input to generate local representation, which is then fed to the transformer-based encoder to output the contextual representation.   
The transformer-based decoder conducts token classiﬁcation for the next word prediction by considering the output of the encoder and predictions of previous tokens. 
It is worth noting that our method can be easily applied to any other encoder-decoder architecture.

\section*{Dual-Path Decoding with Triangular Decomposition Agreement} 

\begin{figure*}[htp]
    \centering 
    \includegraphics[width = 16.5cm]{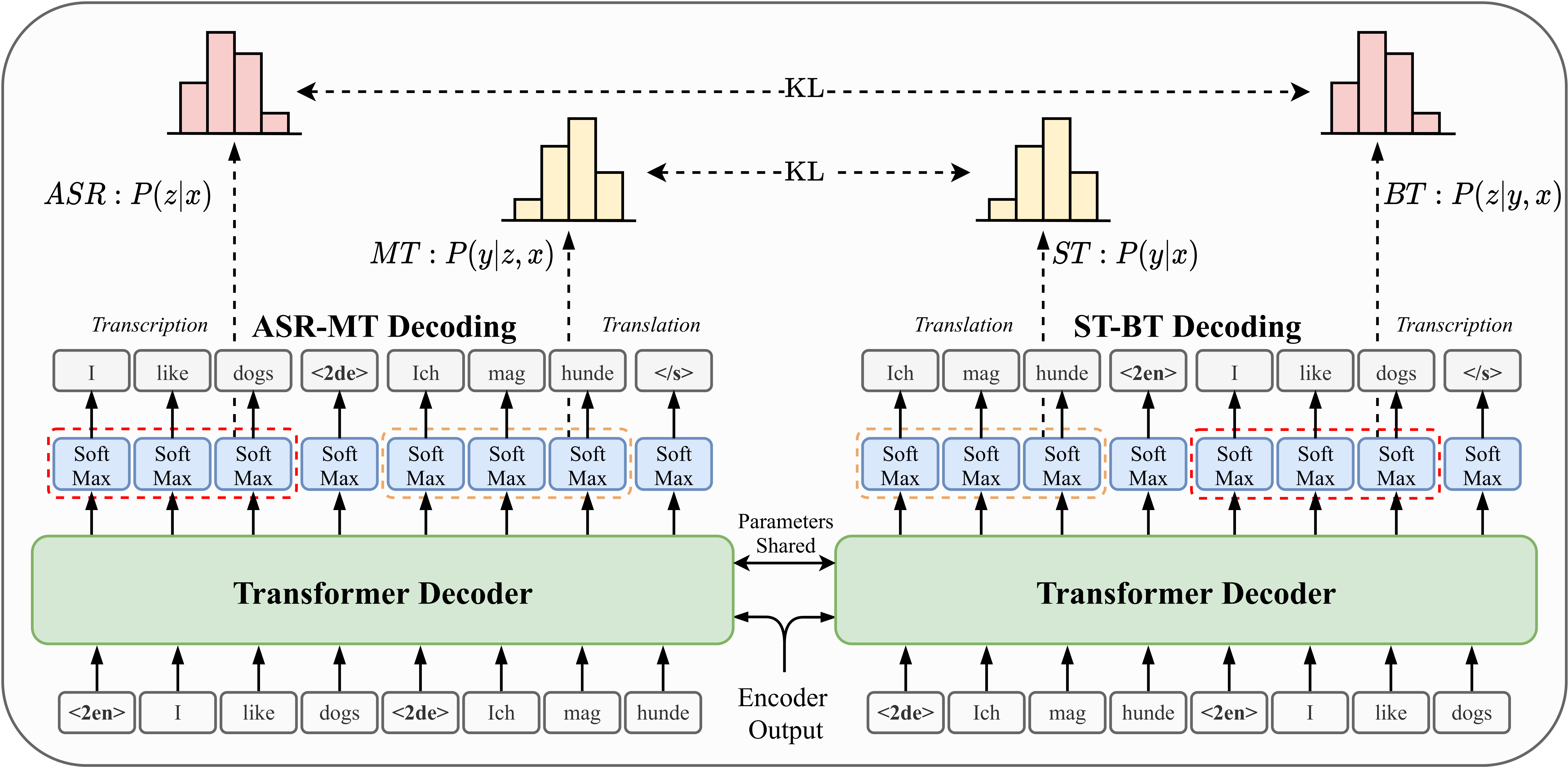}
    \caption{ The overview of dual-path decoding with triangular decomposition agreement.
         The pink histogram is the probability distribution of an example token ``dogs" in the ASR and BT sequences, and similarly, the orange histogram represents the probability distribution of the token ``hunde" in the ST and MT sequences.
        }
    \label{fig:framework}
\end{figure*}

In order to make full use of the triplet data $(\mathbf{x}, \mathbf{z}, \mathbf{y}) \in \mathcal{D}_{ST}$ within a single unified encoder-decoder model, in this work, we directly learn the joint probability $P(\mathbf{y},\mathbf{z} |\mathbf{x)}$ of transcription and translation given the speech input.
To this end, we propose a novel model regularization method called \textbf{T}riangular \textbf{D}ecomposition \textbf{A}greement~(\textbf{TDA}) to fully exploit the association between triplet data, as illustrated in Figure~\ref{fig:framework}.
In this way, the whole training objectives are decomposed into two parts: the standard maximum likelihood of training data, and the regularization terms that indicate the divergence of dual-path based on the current model parameter.
In this section, we start with the dual-path decoding based on the joint probability. 
Furthermore, we introduce additional training objectives in accordance with TDA to improve the agreement of the dual-path decoding.
In the last part, we show the flexibility of our method during inference.

\subsection{Dual-Path Decoding}

We jointly model the generation of transcription and translation in a single decoder. 
In this case, the optimization objective can be calculated by:
\begin{equation}
    \mathcal{L}_{MLE}(\theta)=\sum_{n=1}^{N} \log P\left(\mathbf{y^{(n)}}, \mathbf{z^{(n)}} \mid \mathbf{x^{(n)}} ; \theta\right),
    \label{eq:loss_dual_ways_mle_1}
\end{equation}
where $P\left(\mathbf{y^{(n)}}, \mathbf{z^{(n)}} \mid \mathbf{x^{(n)}} ; \theta\right)$ is the ST model that adopts successive decoding. 
Actually, as shown in Figure~\ref{fig:framework}, there are two different decomposition paths for such conditional probability $P\left(\mathbf{y}, \mathbf{z} \mid \mathbf{x} ; \theta\right)$:
\begin{itemize}
    \item  \textbf{ASR-MT Decoding}. The source transcription $\mathbf{z}$ is first produced by ASR, followed by generating target translation $\mathbf{y}$ through MT:
    \begin{equation}
        \begin{aligned}
            P\left([\mathbf{z},\mathbf{y}] \mid \mathbf{x} ; \theta\right)= P\left( \mathbf{z} \mid \mathbf{x} ; \theta\right) & P\left( \mathbf{y} \mid \mathbf{z}, \mathbf{x} ; \theta\right), \\
        \end{aligned}
        \label{eq:asr_mt_prob}
    \end{equation}
    where $[\mathbf{z},\mathbf{y}]$ means the concatenation of $\mathbf{z}$ and $\mathbf{y}$.
    \item \textbf{ST-BT Decoding}. The decoder output $[\mathbf{y},\mathbf{z}]$ is the concatenation of the translation $\mathbf{y}$ generated by ST and transcription $\mathbf{z}$ obtained by BT process:
    \begin{equation}
        \begin{aligned}
            P\left([\mathbf{y},\mathbf{z}]\mid \mathbf{x} ; \theta\right) =  P\left( \mathbf{y} \mid \mathbf{x} ; \theta\right) & P\left( \mathbf{z} \mid \mathbf{y}, \mathbf{x} ; \theta\right). \\
        \end{aligned}
        \label{eq:st_bt_prob}
    \end{equation}
\end{itemize}
We perform this dual-path decoding in a parameter shared transformer-based decoder and leverage the language tag to distinguish different paths.
Specifically, taking English-to-German ST in Figure~\ref{fig:framework} as an example, ASR-MT decoding utilizes the language tag $\texttt{<2en>}$ as begin-of-sentence (BOS) and generates transcription sequence. 
Unlike the standard decoding, we take $\texttt{<2de>}$ as the end of the transcription sequence.
When the decoder recognizes $\texttt{<2de>}$, it will continue to produce the translation sequence and take end-of-sentence (EOS) as the end mark. 
In this way, we series the transcription-translation sequence through the $\texttt{<2de>}$ identifier. 
Similarly, we adopt $\texttt{<2de>}$ as the BOS to select the ST-BT decoding, while the translation-transcription sequence is concatenated by the $\texttt{<2en>}$.
Therefore, the original training objective in Equation \ref{eq:loss_dual_ways_mle_1} can be rewritten as:
\begin{equation}
    \begin{aligned}
    \mathcal{L}_{MLE}(\theta) = \sum_{n=1}^{N} \log P\left([\mathbf{y^{(n)}}, \mathbf{z^{(n)}}] \mid \mathbf{x^{(n)}} ; \theta\right)  \\ 
     + \sum_{n=1}^{N} \log P\left([\mathbf{z^{(n)}}, \mathbf{y^{(n)}}] \mid \mathbf{x^{(n)}} ; \theta\right).
    \end{aligned}
    \label{eq:loss_dual_ways_mle}
\end{equation}

\subsection*{Model Regularization}
According to the chain decomposition rule, output probabilities of these dual paths should be identical if the learned model is perfect (we drop $\theta$ for concise):
  \begin{equation}
        \begin{aligned}
            P\left([\mathbf{y},\mathbf{z}]\mid \mathbf{x}\right) & =  P\left( \mathbf{y} \mid \mathbf{x}\right) P\left( \mathbf{z} \mid \mathbf{y}, \mathbf{x}\right) \\
            & = P\left( \mathbf{y} \mid \mathbf{z}, \mathbf{x} \right) P\left( \mathbf{z} \mid \mathbf{x} \right) = P\left([\mathbf{z},\mathbf{y}] \mid \mathbf{x} \right).
        \end{aligned}
        \label{eq:agreement}
    \end{equation}
However, if these two paths are optimized independently by MLE (like Equation~\ref{eq:loss_dual_ways_mle}), there is no guarantee that the above equation will hold. 
To handle this problem, we introduce two word-level Kullback-Leibler~(KL) divergence based on the output probability as the regularization terms, aiming to enhance the agreement between ASR-MT and ST-BT decoding paths.
Since KL divergence is asymmetric, it involves two directions:
\begin{equation}
    \begin{aligned}
        \overrightarrow{\text{KL}} & = \text{KL} \left( P\left( \mathbf{y} \mid \mathbf{z}, \mathbf{x} \right) \mid\mid P\left( \mathbf{y} \mid \mathbf{x} \right) \right) \\ 
        & + \text{KL} \left(P\left( \mathbf{z} \mid \mathbf{x} \right) \mid\mid P\left( \mathbf{z} \mid \mathbf{y}, \mathbf{x} \right) \right)\\
        & = \sum_{t=1}^{|\mathbf{y}|}  P\left(y_{t} \mid y_{<t},z_{\leq|\mathbf{z}|}, \mathbf{x} \right) \log \frac{P\left(y_{t} \mid y_{<t},z_{\leq|\mathbf{z}|}, \mathbf{x} \right)}{P\left(y_{t} \mid y_{<t}, \mathbf{x} \right)} \\
        & + \sum_{t'=1}^{|\mathbf{z}|}  P\left(z_{t'} \mid z_{<t'}, \mathbf{x} \right) \log \frac{P\left(z_{t'} \mid z_{<t'}, \mathbf{x} \right)}{P\left(z_{t'} \mid z_{<t'},y_{\leq|\mathbf{y}|}, \mathbf{x} \right)}, \\
        \overleftarrow{\text{KL}} & =   \text{KL} \left( P\left( \mathbf{y} \mid \mathbf{x} \right) \mid\mid P\left( \mathbf{y} \mid \mathbf{z}, \mathbf{x} \right) \right) \\ 
        & + \text{KL} \left( P\left( \mathbf{z} \mid \mathbf{y}, \mathbf{x} \right) \mid\mid P\left( \mathbf{z} \mid \mathbf{x} \right) \right) \\
        & = \sum_{t'=1}^{|\mathbf{y}|}  P\left(y_{t'} \mid y_{<t'}, \mathbf{x} \right) \log \frac{P\left(y_{t} \mid y_{<t'}, \mathbf{x} \right)}{P\left(y_{t'} \mid y_{<t'},z_{\leq|\mathbf{z}|}, \mathbf{x} \right)} \\
        & + \sum_{t=1}^{|\mathbf{z}|}  P\left(z_{t} \mid z_{<t},y_{\leq|\mathbf{y}|}, \mathbf{x} \right) \log \frac{P\left(z_{t} \mid z_{<t},y_{\leq|\mathbf{y}|}, \mathbf{x} \right)}{P\left(z_{t} \mid z_{<t}, \mathbf{x} \right)}
    \end{aligned}
    \label{eq:kl2}
\end{equation}
where $|\mathbf{y}|$ and $|\mathbf{z}|$ represent the length of translation and transcription respectively.
The entire regularization term is summarized as: 
\begin{equation}
    \begin{aligned}
        \mathcal{L}_{TDA}(\theta)= \overrightarrow{\text{KL}} +  \overleftarrow{\text{KL}}.
    \end{aligned}
    \label{eq:loss_tda}
\end{equation}
The Equation~\ref{eq:agreement} holds when these regularization terms are $0$, otherwise regularization terms will guide the training process to reduce the disagreement of output probabilities of dual-path decoding.
Besides, we assign a weighting term to this loss and combine it with the MLE loss to obtain the entire model training objective, as described by:
\begin{equation} 
    \mathcal{L}_{}(\theta) = \mathcal{L}_{MLE}(\theta) - \lambda \mathcal{L}_{TDA}(\theta),
    \label{eq:loss_overall}
\end{equation}
where $\lambda$ is a hyper-parameter to balance the preference between the ground truth and agreement distribution.

\subsection*{Inference}
Since we can adapt language tags to switch the ASR-MT and ST-BT decoding paths, it gives the flexibility of the inference strategy.
The well-trained model is naturally transformed as the ASR and E2E-ST models by choosing ASR-MT and ST-BT decoding paths, respectively.
Specifically, we directly select the ST-BT path to conduct the English-to-German ST task, and terminate the inference when generating language identifier $\texttt{<2en>}$.
In this way, our proposed approach maintains the same decoding speed as the traditional E2E-ST model. 
Similarly, we can gain the ASR result by selecting the ASR-MT decoding path and terminating the decoding when $\texttt{<2de>}$ is recognized. 
In addition, our approach can simultaneously leverage two ways to inference and adopt the way of premature termination for different scenarios, that is, only the corresponding text is displayed, or both transcription and translation are displayed to the user at the same time.

\section{Experiments}

\subsection*{Setup}
 
We consider restricted and extended settings on the benchmark MuST-C to evaluate the effectiveness of our proposed approach. 
For the restricted setting, we run experiments on the MuST-C dataset with all 8 languages.
For comparison in practical scenarios, we extend the above setting to verify the gain of our method on English-to-German and English-to-French translation directions with available external ASR and MT data.  

\subsubsection{MuST-C Dataset.} 
MuST-C~\cite{Gangi2019MuSTCAM} is a publicly large-scale multilingual speech-to-text translation corpus, consisting of triplet data sources: source speech, source transcription, and target translation.
The speech sources of MuST-C are from English TED Talks, which are aligned at the sentence level with their manual transcriptions and translations. 
MuST-C contains translations from English (EN) to 8 languages: Dutch (NL), French (FR), German (DE), Italian (IT), Portuguese (PT), Romanian (RO), Russian (RU), and Spanish (ES).
The statistics of different language pairs are illustrated in Table \ref{tab:must-c_stat}.

\begin{table}[t]
    \centering
    \scalebox{0.92}{
    \begin{tabular}{l|cccccc} 
        \hline
        \toprule
        \multirow{2}*{Language} & Sentence  & Speech    & Source    & Target \\
                                & Pair      & Duration  & Words     & Words \\
        \midrule
        German~(DE)     & 234$\mathrm{~K}$ & 408$\mathrm{~hrs}$ & 4.3$\mathrm{~M}$ & 4.0$\mathrm{~M}$ \\
        French~(FR)     & 280$\mathrm{~K}$ & 492$\mathrm{~hrs}$ & 5.2$\mathrm{~M}$ & 5.4$\mathrm{~M}$ \\
        Spanish~(ES)    & 270$\mathrm{~K}$ & 504$\mathrm{~hrs}$ & 5.3$\mathrm{~M}$ & 5.1$\mathrm{~M}$ \\
        Italian~(IT)    & 258$\mathrm{~K}$ & 465$\mathrm{~hrs}$ & 4.9$\mathrm{~M}$ & 4.6$\mathrm{~M}$ \\
        Dutch~(NL)      & 253$\mathrm{~K}$ & 442$\mathrm{~hrs}$ & 4.7$\mathrm{~M}$ & 4.3$\mathrm{~M}$ \\
        Portuguese~(PT) & 211$\mathrm{~K}$ & 385$\mathrm{~hrs}$ & 4.0$\mathrm{~M}$ & 3.8$\mathrm{~M}$ \\
        Romanian~(RO)   & 240$\mathrm{~K}$ & 432$\mathrm{~hrs}$ & 4.6$\mathrm{~M}$ & 4.3$\mathrm{~M}$ \\
        Russian~(RU)    & 270$\mathrm{~K}$ & 489$\mathrm{~hrs}$ & 5.1$\mathrm{~M}$ & 4.3$\mathrm{~M}$ \\
        \bottomrule
        \hline
    \end{tabular}
    }
    \caption{The statistics of 8 translation directions in the MuST-C dataset.}
    \label{tab:must-c_stat}
\end{table}

\subsubsection*{External ASR and MT Datasets.}
We introduce the LibriSpeech dataset \cite{Panayotov2015LibrispeechAA} as the external ASR data. 
The LibriSpeech ASR dataset is derived from audiobooks that are part of the LibriVox project.
This dataset contains 960 hours of speech samples in English and approximately 290K speech-transcription pair samples, in which the transcription texts are not punctuated and capitalized. 
We adopt English-to-German and English-to-French WMT14~\cite{Bojar2014FindingsOT} training data as the external MT parallel corpus in the extended setting, which consists of 4M and 30M bilingual sentence pairs, respectively. 

\subsubsection*{Pre-processing of Data.} 
We follow \textsc{Fairseq} S2T~\cite{Wang2020FairseqSF} recipes to perform data pre-processing. 
For speech data, both in LibriSpeech and MuST-C, acoustic features are 80-dimensional log-mel filter banks extracted with a stepsize of 10ms and a window size of 25ms. 
The acoustic features are normalized by global channel mean and variance. 
In addition, the SpecAugment method \cite{Park2019SpecAugmentAS} is applied for all experiments, and the samples of more than 3000 frames are removed. 
As for text data in MuST-C and WMT14, we reserve punctuation, as well as the original word splitting and normalization. 
We lowercase all transcription sentences in the Librispeech ASR dataset, capitalize the first letter of all sentences and put a full stop at the end of the sentence to be consistent with MuST-C and WMT14 datasets.
For sub-wording, we employ the unigram sentencepiece\footnote{https://github.com/google/sentencepiece} model to build a sub-word vocabulary with a size of 10000.
On each translation direction, the sentencepiece model is learned on text data from the training set, and the dictionary is shared across source and target languages.

\begin{table*}[htb]
    \small
    \normalsize
    \centering
    \setlength{\tabcolsep}{1.5mm}{
    \scalebox{0.95}{
        \begin{tabular}{l|c|c|cccccccc|l}
            \hline
            \toprule
            \textbf{Model}  & \textbf{Params.} & \textbf{Extra.} & \textbf{EN-DE} & \textbf{EN-FR} & \textbf{EN-RU} & \textbf{EN-ES} & \textbf{EN-IT} & \textbf{EN-RO} & \textbf{EN-PT} & \textbf{EN-NL} & \textbf{Avg.} \\ \midrule
            ESPnet ST     & 31M  & $\times$      & 22.9  & 32.8  & 15.8  & 28.0 & 23.8  & 21.9  & 28.0  & 27.4  & 25.1 \\
            ESPnet Cascaded & 84M & $\times$      & 23.6 & 33.8 & 16.4 & 28.7 & 24.0 & 22.7 & 29.0 & 27.9 & 25.8 \\ 
            Fairseq ST            & 31M & $\times$      & 22.7 & 32.9 & 15.3 & 27.2 & 22.7 & 21.9 & 28.1 & 27.3 & 24.8 \\ 
            Fairseq Multi-ST     & 76M & $\times$      & 24.5 & 34.9 & 16.0 & 28.2 & 24.6 & 23.8 & \textbf{31.1} & 28.6 & 26.5 \\ 
            AFS                 &  -  & $\times$      & 22.4 & 31.6 & 14.7 & 26.9 & 23.0 & 21.0 & 26.3 & 24.9 & 23.9 \\
            Dual-Decoder   & 48M & $\times$      & 23.6 & 33.5 & 15.2 & 28.1 & 24.2 & 22.9 & 30.0 & 27.6 & 25.7 \\ 
            W2V2-Transformer    & -  & \checkmark    & 22.3 & 34.3 & 15.8 & 28.7 & 24.2 & 22.4 & 29.3 & 28.2 & 25.6  \\ 
            LNA-E,D            &  76M   & $\times$    & 24.3 & 34.6 & 15.9 & 28.4 & 24.4 & 23.3 & 30.5 & 28.3 & 26.2  \\ 
            Adapter Tuning      &  76M  & $\times$    & 24.6 & 34.7 & \textbf{16.4} & 28.7 & 25.0 & 23.7 & 31.0 & 28.8 & 26.6 \\ 
            \midrule
            E2E-ST-$\text{Base}^{s}$  & 31M  & $\times$      & 22.8 & 33.0 & 15.2 & 27.2 & 22.9 & 21.6 & 28.0 & 27.3  & 24.8 \\ 
            E2E-ST-$\text{JT}^{s}$    &  32M   & $\times$      & 23.1 & 32.8 & 14.9 & 27.5 & 23.6 & 22.1 & 28.7 & 27.8 & 25.1  \\ 
            E2E-ST-$\text{TDA}^{s}$     &   32M       & $\times$      & 24.3 & 34.6 & 15.9 & 28.3 & 24.2 & 23.4 & 30.3 & 28.7 & 26.2  \\ 
            \midrule
            E2E-ST-$\text{Base}^{m}$  & 74M  & $\times$      & 23.5 & 33.8 & 15.5 & 27.8 & 23.4 & 22.8 & 28.6 & 27.5 & 25.4  \\ 
            E2E-ST-$\text{JT}^{m}$   &  76M     & $\times$      & 23.2 & 34.1 & 14.9 & 28.2 & 23.1 & 22.4 & 28.4 & 27.9 & 25.3  \\ 
            E2E-ST-$\text{TDA}^{m}$      &   76M      & $\times$ & \textbf{25.4} & \textbf{36.1} & \textbf{16.4} & \textbf{29.6} & \textbf{25.1} & \textbf{23.9} & \textbf{31.1}           & \textbf{29.6} & \textbf{27.2}     \\
            \bottomrule
            \hline
        \end{tabular}}}
    \caption{BLEU scores of different methods on MuST-C tst-COMMON set. ``Extra." indicates whether the method uses additional data. ``Params." represents the parameter scale of the model. The superscripts $s$ and $m$ represent the small model and medium model, respectively.}
    \label{tab:main_results_bleu}
\end{table*}

\begin{table*}[htb]
    \small
    \normalsize  
    \centering
    \setlength{\tabcolsep}{1.5mm}{
        \scalebox{0.9}{
            \begin{tabular}{l|cccccccc|c}
                \hline    
                \toprule
                \textbf{Model} &  \textbf{EN-DE} & \textbf{EN-FR} & \textbf{EN-RU} & \textbf{EN-ES}  &\textbf{EN-IT} &\textbf{EN-RO} & \textbf{EN-PT} & \textbf{EN-NL} & \textbf{Avg.}      \\
                \midrule
                E2E-ST-$\text{Base}^{s}$   &18.2   &17.2   &17.7   &17.7   &17.9   &18.1   &19.1   &17.6   &17.9 \\
                E2E-ST-$\text{JT}^{s}$         &17.2   &16.7   &16.9   &16.4   &16.8   &16.8   &17.8   &16.6   &16.9 \\
                E2E-ST-$\text{TDA}^{s}$                &16.4   &15.6   &16.6   &16.4   &16.2   &16.6   &16.9   &16.2   &16.4 \\
                \midrule
                E2E-ST-$\text{Base}^{m}$     &16.8   &16.9   &16.9   &16.9   &17.0   &17.0   &17.4   &16.7   &17.0 \\
                E2E-ST-$\text{JT}^{m}$        &16.3   &15.2   &16.5   &15.1   &16.3   &16.9   &16.8   &15.6   &16.1 \\
                E2E-ST-$\text{TDA}^{m}$  &\textbf{14.9} &\textbf{14.1} &\textbf{15.7} &\textbf{14.4} &\textbf{15.2} &\textbf{15.4} &\textbf{16.5} &\textbf{14.9} &\textbf{15.1} \\ 
                \bottomrule
                \hline
            \end{tabular}
        }
    }  
    \caption{WER scores of different methods on Must-C tst-COMMON set.}
    \label{tab:main_results_wer}
\end{table*}

\subsubsection{Methods.}
We compare our proposed approach (E2E-ST-TDA) with several baseline methods in the experiment:
\begin{itemize}
    \item E2E-ST-Base: we optimize the E2E-ST model with the training process proposed in \citet{Wang2020FairseqSF}. 
    The model is first pre-trained with speech-transcription pairs and then directly fine-tuned by speech-translation pairs. 
    \item E2E-ST-JT: we train the E2E-ST model with multi-task learning, including ASR and ST tasks.
    \item E2E-ST-TDA: we extend the E2E-ST-Base method with the proposed model regularization method TDA. 
    
\end{itemize}
We implement these methods with small and medium model sizes respectively, in which we adopt superscripts $s$ and $m$ to represent the correspondent model size.
Besides, we also compare E2E-ST-TDA with other E2E-ST baselines, which include using only MuST-C data and using external data: 
ESPnet ST and Cascaded~\cite{Inaguma2020ESPnetSTAS}, Fairseq ST and Multi-ST~\cite{Wang2020FairseqSF}, AFS~\cite{Zhang2020AdaptiveFS}, Dual-Decoder~\cite{Le2020DualdecoderTF}, W2V2-Transformer~\cite{Han2021LearningSS}, LNA-E,D~\cite{Li2021MultilingualST}, Adapter Tuning~\cite{Le2021LightweightAT} and Chimera~\cite{Han2021LearningSS}.

\subsubsection*{Training Details and Evaluation.}
All experiments are implemented based on the \textsc{Fairseq}\footnote{https://github.com/pytorch/fairseq}~\cite{Ott2019fairseqAF} toolkit. 
We adopt the transformer-based backbone for all models, consisting of 2 layers of one-dimensional convolutional layers with a down-sampling factor of 4, 12 Transformer encoder layers, and 6 Transformer decoder layers. 
More specifically, for the small model, we set the size of the self-attention layer, the feed-forward network, and the head to 256, 2048, and 4, respectively; for the medium model, the above parameters are set to 512, 2048, and 8, respectively. 
All models are initialized using the pre-trained ASR speech encoder to speed up the model convergence.
During training, we use the adam optimizer~\cite{Kingma2015AdamAM} with a learning rate set to 0.002 to update model parameters with 10K warm-up updates.
The label smoothing and dropout ratios are set to 0.1 and 0.3, respectively. 
In practice, we train all models with 2 Nvidia Tesla-V100 GPUs and it takes 1-2 days to finish the whole training.
The batch size in each GPU is set to 10000, and we accumulate the gradient for every 4 batches. 
During inference, we average the model parameters on the 10 best checkpoints based on the performance of the MuST-C dev set, and adopt beam search strategy with beam size of 5.
In our experiments, we report the WER score for ASR task and the case-sensitive BLEU score \cite{Papineni2002BleuAM} for ST task using sacreBLEU\footnote{https://github.com/mjpost/sacrebleu, with a configuration
of 13a tokenizer, case-sensitiveness, and full punctuation}.

\subsection{Main Results}
\subsubsection*{E2E-ST Performance on MuST-C.} 
We evaluate the E2E-ST performance of our proposed method on the MuST-C dataset with 8 languages.
As illustrated in Table~\ref{tab:main_results_bleu}, we can observe that our approach E2E-ST-TDA significantly outperforms two baselines E2E-ST-Base and E2E-ST-JT, in all languages.
More specifically, E2E-ST-TDA obtains an average BLEU score improvement of 1.4/1.8 respectively compared to E2E-ST-Base with different model sizes.
These results demonstrate that our approach leverages triangular decomposition agreement to fully exploit the triplet training data, leading to better translation performance. 
In addition, we include the results from previous work, such as ESPnet ST and Cascaded, Fairseq ST and Multi-ST, AFS, Dual-Decoder, W2V2-Transformer, LNA-E,D and Adapter Tuning.
We can find that our implemented baseline E2E-ST-$\text{Base}^{s}$ achieves similar performance as ESPnet ST and Fairseq ST, while our proposed method outperforms the cascaded system - ESPnet Cascaded trained on the same data.
Compared with Dual-Decoder, our proposed method E2E-ST-$\text{TDA}^{m}$ gains more remarkable improvement in all languages.
Different from the Dual-Decoder that considers the interaction between loosely coupled ASR decoder and ST decoder, our method leverages the agreement of dual-paths with a shared decoder and saves the inference time.
Besides, our approach outperforms some methods that use external data and multilingual versions.
Instead of log-mel filter banks, W2V2-Transformer adopts pre-trained wav2vec 2.0~\cite{Baevski2020wav2vec2A} to improve the E2E-ST performance.
Multilingual ST models, including Fairseq Multi-ST, LNA-E,D, and Adapting Tuning, beat most baselines since the target languages are mostly Indo-European languages with similar grammatical structures, which better help learn shared model parameters.
As we can see, our proposed approach achieves state-of-the-art performance on all translation directions among all cascade and end-to-end systems in Table~\ref{tab:main_results_bleu}, which proves the effectiveness of our method.

\subsubsection*{ASR Performance on MuST-C.} 
Since our method involves the ASR model during training, we also compare correspondent performance with two baselines: E2E-ST-Base and E2E-ST-JT.
E2E-ST-Base denotes the performance of the pre-trained ASR model used to initialize all E2E-ST models.  
As illustrated in Table~\ref{tab:main_results_wer}, our approach can significantly improve the performance on ASR tasks, reducing 1.5/1.9 WER scores over E2E-ST-Base with the small and medium model ,respectively.
The performance improvement indicates that our proposed method can make full use of the entire training data by achieving better agreement in the training process to jointly improve the performance on E2E-ST and ASR tasks.

\begin{table}[t]
    \normalsize  
    \centering
    \setlength{\tabcolsep}{1.5mm}{
        \scalebox{0.92}{
            \begin{tabular}{l|c|cc|cccc}
                \hline    
                \toprule
                \multicolumn{1}{c|}{\textbf{Model}} &  \textbf{Params.} &  \textbf{EN-DE} & \textbf{EN-FR} & \textbf{Avg.}  \\

                \midrule
                MT                      & -     & 32.2  & 46.1  & 39.2 \\
                Cascaded ST             & -     & 27.0  & \textbf{38.6}  & \textbf{32.8}  \\
                 \midrule
                Chimera Mem-16        & 165M  & 25.6  & 35.0  & 30.3 \\
                Chimera                 & 165M  & 26.3  & 35.6  & 31.0 \\
                \midrule
                E2E-ST-$\text{Base}^{m}$    & 74M   & 25.8  & 35.9  & 30.9  \\
                E2E-ST-$\text{TDA}^{m}$               & 76M   & \textbf{27.1} & 37.4 & 32.3   \\
                \bottomrule
                \hline
            \end{tabular}
        }
    }  
    \caption{BLEU scores on MuST-C tst-COMMON set in an extended setting. The external data includes 960 hours of LibriSpeech ASR data and WMT14 EN-DE/FR MT data.}
    \label{tab:external_data_results}
\end{table}

\subsubsection*{E2E-ST Performance on Extended Setting.}
We further verify the effectiveness of our proposed method with external ASR and MT data.
Table~\ref{tab:external_data_results} shows the results of all methods, including Cascaded ST, E2E-ST-Base, E2E-ST-TDA, and the recent SOTA method Chimera.
For E2E-ST-Base and E2E-ST-TDA, we first train two MT models with the mixed data of WMT14 and MuST-C on EN-DE/EN-FR translation directions and then translate the transcriptions in Librispeech to build the additional triplet corpus.
We also pre-train the ASR model on the mixed data of Librispeech and MuST-C to initialize all E2E-ST models.
These ASR and MT models are used to perform Cascaded ST.
The first two rows in the Table~\ref{tab:external_data_results} show that the translation quality drops sharply when the output of the ASR model is fed as the input of the MT model compared with the clean transcription input. 
Our approach E2E-ST-$\text{TDA}^{m}$ significantly surpasses E2E-ST-Base and Chimera under this setting, achieving a smaller parameter scale than Chimera at the same time. 
These experimental results prove that our proposed method can stably improve the translation quality of the E2E-ST system even with external data.

\begin{table}[t]
    \small
    \normalsize  
    \centering
    \setlength{\tabcolsep}{1.0mm}{
        \scalebox{0.9}{
            \begin{tabular}{l|ccc|cccccccc}
                \hline    
                \toprule
                \multirow{2}*{\textbf{Model}}  & \multicolumn{3}{c|}{\textbf{BLEU}}  & \multicolumn{3}{c}{\textbf{WER}} \\
                                            & EN-DE   & EN-FR   & Avg.  & EN-DE     & EN-FR     & Avg.   \\
                \midrule
                 E2E-ST-$\text{Base}^{m}$   & 23.5    & 33.8    & 28.7  & 16.8      & 16.9      & 16.9  \\
                 \ \ \ + WordKD              & 23.9    & 34.5    & 29.2  & -         & -         & -     \\
                 \ \ \ + SeqKD              & 24.5    & 35.2    & 29.9  & -         & -         & -     \\
                 E2E-ST-$\text{TDA}^{m}$    & \textbf{25.4}    & \textbf{36.1}    & \textbf{30.8}  & \textbf{14.9}      & \textbf{14.1}      & \textbf{14.5}  \\
                 \ \ \ w/o KL               & 23.8    & 34.5    & 29.2  & 16.3      & 16.0      & 16.2  \\
                \bottomrule
                \hline
            \end{tabular}
        }
    }  
    \caption{BLEU and WER scores of ablation study on MuST-C tst-COMMON set. ``w/o" means without.}
    \label{tab:ablation_study}
\end{table}

\subsection{Analysis}
\subsubsection*{Ablation Study.}

In order to analyze the effectiveness of different modules in our method, we carry out an ablation study on EN-DE and EN-FR translation directions in the MuST-C dataset.
As shown in Table~\ref{tab:ablation_study}, besides E2E-ST-Base and E2E-ST-TDA, we evaluate the performance of three models: E2E-ST-Base with word-level or sequence-level knowledge distillation (WordKD/SeqKD)~\cite{Kim2016SequenceLevelKD} and E2E-ST-TDA without KL regularization terms.
Actually, E2E-ST-TDA can be considered as the expansion of WordKD and SeqKD methods, while these approaches merely reduce the mismatch between MT and E2E-ST models.
E2E-ST-TDA yields better translation results than E2E-ST-Base + WordKD/SeqKD, since our method can fully leverage the triangular relationship in training data.
On the other hand, compared with E2E-ST-Base, the performance improvement of E2E-ST-TDA without KL regularization terms seems marginal.
It indicates that optimizing the model with only MLE loss fails to fully utilize the association between triplet data. 

\subsubsection{Effect of Model Size.}

\begin{figure}[t]
    \centering
    \includegraphics[width=1\columnwidth]{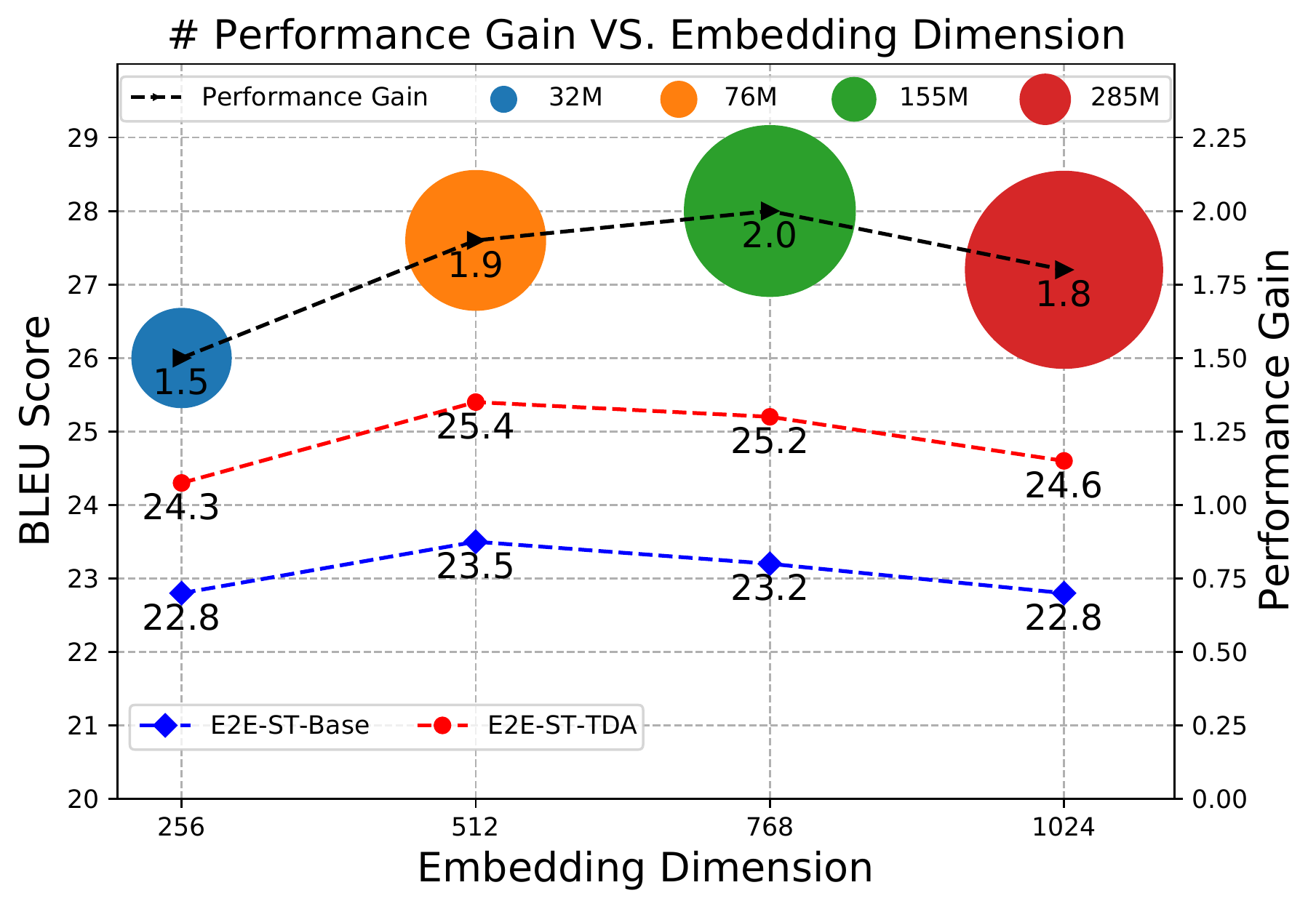} 
    \caption{The impact of model size on performance. The area of the circle represents the parameter size of the model.}
    \label{fig:gain_vs_dim}
\end{figure}

As illustrated in Table~\ref{tab:main_results_bleu}, the bigger model seems to obtain better improvement when using our method.
To further verify the performance of our method with different model sizes, we conduct experiments on the MuST-C EN-DE dataset.
We adopt dimensions ranging from $(256, 512, 768, 1024)$ for quick experiments.
The detailed results are shown in Figure~\ref{fig:gain_vs_dim}.
From the figure, we can see that with the increase of the embedding dimension, the performance gain increases first and then remains stable,
while the model performance of both E2E-ST-Base and E2E-ST-TDA increase first and then decrease. 
It is because that a larger model (with a higher embedding dimension) typically requires more data for training, suffering from the overfitting risk and decreased efficiency.

\subsubsection{Effect of Hyper-parameter $\lambda$.}
\begin{figure}[t]
    \centering
    \includegraphics[width=1\columnwidth]{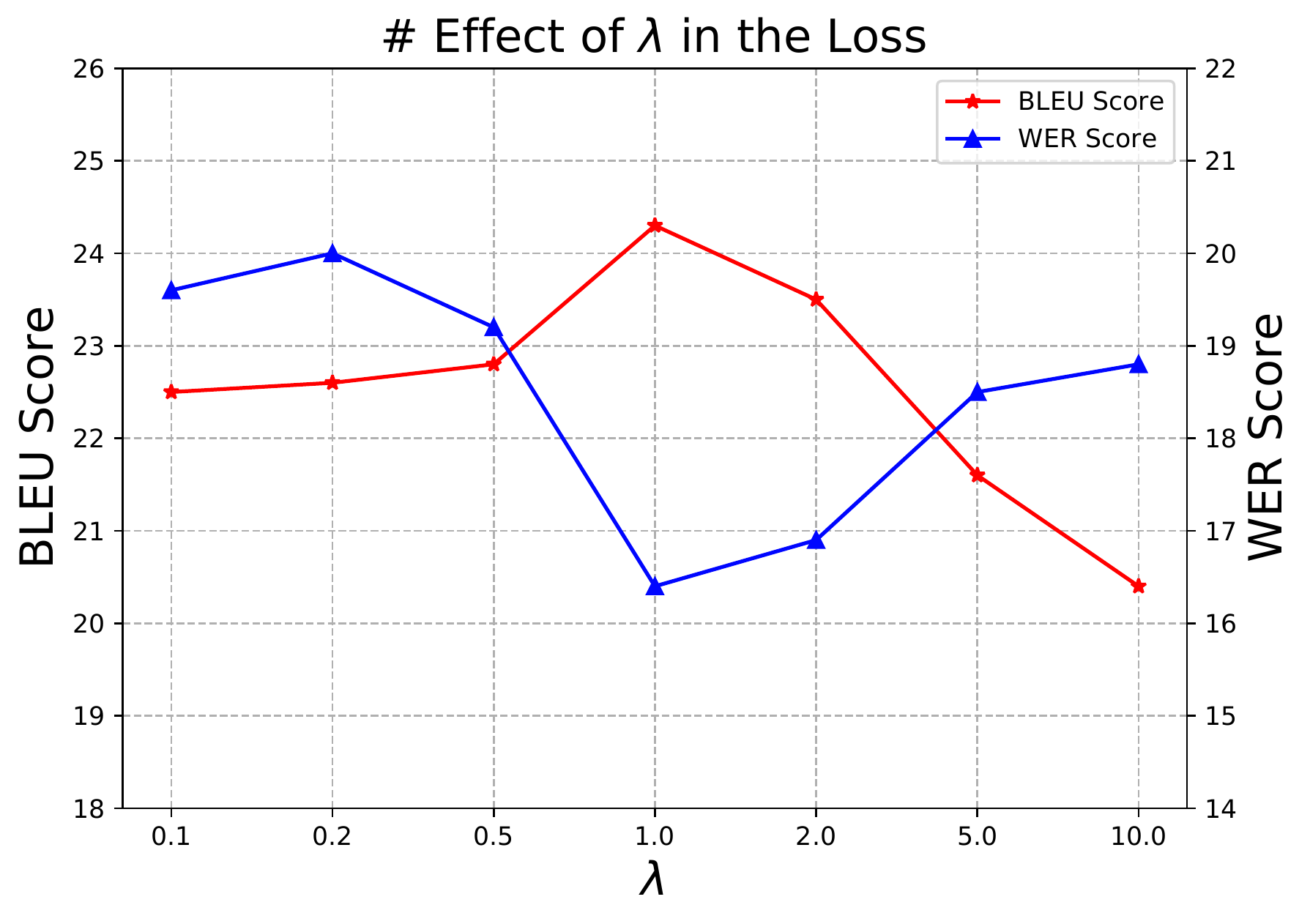} 
    \caption{The impact of the hyper-parameter $\lambda$ in the loss function on performance.}
    \label{fig:effect_of_lambda}
\end{figure}

In our experiments, we attempt different settings ($\lambda = 0.1, 0.5, 1.0, 2.0, 5.0, 10.0$), and find that $\lambda = 1.0$ achieves the best BLEU and WER scores on MuST-C EN-DE development set. 
These results are shown in Figure~\ref{fig:effect_of_lambda}. 
A larger $\lambda$ will make the model pay more attention to triangular decomposition agreement, while a smaller $\lambda$ will make the model pay more attention to optimizing the dual-path. 
In the early stage of model training, since the learned parameters are not perfect, a larger $\lambda$ will cause the parameters of the model to be constrained in the wrong position by agreement earlier, making it challenging to further optimize the model. 
However, a smaller $\lambda$ will make the dual-path too independent, and it is not easy to narrow the output representation.

\section{Related Work}

\subsubsection*{Speech Translation.} 
Early ST methods~\cite{Ney1999SpeechTC,Matusov2005OnTI,Sperber2017TowardRN, Cheng2018TowardsRN} cascade the ASR subsystem and the MT subsystem. 
With the rapid development of deep learning, the neural networks widely used in ASR and MT have been adapted to construct a new end-to-end speech-to-text translation paradigm.
However, due to the scarcity of triplet training data, developing an E2E-ST model that does not use intermediate transcription is still very challenging. 
Thus, various techniques have been proposed to ease the training process by using source transcriptions, including pre-training~\cite{Bansal2019PretrainingOH, Wang2020CurriculumPF, Wang2020BridgingTG}, multi-task learning~\cite{Weiss2017SequencetoSequenceMC,Anastasopoulos2018TiedML,Sperber2019AttentionPassingMF}, meta-learning~\cite{Indurthi2020EndendST}, interactive decoding~\cite{Liu2020SynchronousSR}, consecutive decoding~\cite{Dong2021ConsecutiveDF}, and adapter tuning~\cite{Le2021LightweightAT}. 
Among them, the pre-training strategy is a simple and effective way that pre-train different components of the ST system and merges them into one. 
However, the training process of these methods only loosely couples the two type two-tuple data. It does not fully explore the potential connections between the triplet data. Therefore, how to fully mining the associations between the scarce triplet data still remains a crucial issue to improve the performance of E2E-ST.
Following this research line, we introduce TDA to fully exploit the association relationship in triplet data.

\subsubsection*{Agreement Regularization.}
One line of agreement regularization attempts to regularize model predictions to be invariant with minute perturbations on input data, which focused on semi-supervised learning areas.
The minute perturbations can be random noise~\cite[]{Zheng2016ImprovingTR}, adversarial noise~\cite[]{Miyato2019VirtualAT,Carmon2019UnlabeledDI,Zhu2020FreeLBEA,Jiang2020SMARTRA}, gaussian noise~\cite[]{Aghajanyan2021BetterFB} and various data augmentation methods~\cite[]{Ye2019UnsupervisedEL,Xie2020UnsupervisedDA}. 
Another line tries to take into consideration the agreement between the different models, especially in sequence modeling. 
For instance, there are some attempts in speech recognition~\cite{Mimura2018ForwardBackwardAD}, neural machine translation~\cite[]{Liu2016AgreementOT,Zhang2019RegularizingNM}, and speech synthesis~\cite[]{Zheng2019ForwardBackwardDF}, which try to improve the performance by integrating the predicted probability from forward and backward decoding sequences. 
Our method is most similar to the latter, but we aim to constrain the agreement between the probability distributions of two directions sequences in a single unified model. 

\section{Conclusion}
In this paper, we propose a flexible and effective regularization method for the ST task, namely Triangular Decomposition Agreement~(TDA), which relies on the agreement between inherent and unexplored dual decomposition paths ASR-MT and ST-BT of ST. 
In our method, two Kullback-Leibler divergences are added to the standard training objective as regularization terms to resolve the mismatch between the joint probability distributions of dual-path ASR-MT and ST-BT. 
In addition, our approach can use two paths for inference and adopt early termination methods for different scenarios to ensure efficient inference speed. 
Empirical evaluations of the eight translation directions of MuST-C demonstrate that our proposed approach leads to significant improvements compared with strong baseline systems.

\section*{Acknowledgements}
This research is supported by grants from the National Natural Science Foundation of China (Grant No.62072423) and National Major Scientific Instruments and Equipments Development Project (Grant No.61727809). This work is also supported by Alibaba Innovative Research Program. We appreciate Dongqi Wang, Linan Yue, Dexin Wang and Linlin Zhang for the fruitful discussions. We thank the anonymous reviewers for helpful feedback on early versions of this work. This work is done during the first
author’s internship at Alibaba DAMO Academy.

\bibliography{ref_overleaf}
\end{document}